\documentclass[twoside,12pt]{article}
\usepackage{jmlr2e}
\usepackage[12pt]{moresize}

\usepackage{microtype}
\usepackage{graphicx}
\usepackage{subfigure}
\usepackage{booktabs,multirow} 
\usepackage{amsmath}
\usepackage{amssymb}
\usepackage{multirow}
\usepackage{hyperref}

\usepackage{algorithm,algorithmic,dblfloatfix}

\title{Learning the Multiple Traveling Salesmen Problem\\  with Permutation Invariant Pooling Networks}

\author{\name Yoav Kaempfer \email yoavkaempfer@mail.tau.ac.il\\ 
       \addr The School of Computer Science \\
       Tel Aviv University \\
	   Tel Aviv, Israel \\
       \AND 
       \name Lior Wolf \email wolf@cs.tau.ac.il \\
       \addr Facebook AI Research \& \\
       The School of Computer Science \\
       Tel Aviv University \\
	   Tel Aviv, Israel \\}

\begin{document}
\maketitle

	\begin{abstract}
While there are optimal TSP solvers, as well as recent learning-based approaches, the generalization of the TSP to the Multiple Traveling Salesmen Problem is much less studied. Here, we design a neural network solution that treats the salesmen, cities and depot as three different sets of varying cardinalities. We apply a novel technique that combines elements from recent architectures that were developed for sets, as well as elements from graph networks. Coupled with new constraint enforcing output layers, a dedicated loss, and a search method, our solution is shown to outperform all the meta-heuristics of the leading solver in the field.
	\end{abstract}
	
	\section{Introduction}
	
	One of the most fundamental goals of machine learning is to provide learned approximations to combinatorial optimization problems of intractable complexities. Image segmentation, pose estimation, sequence  alignment, parsing, machine translation, protein design, etc. could be defined as large-scale combinatorial problems, in which variables are assigned discrete values, based on a cost term that involves the observations and perhaps elements of the training data.
	
	Classical NP hard combinatorial problems pose a unique set of challenges. First, obtaining a training sample often requires solving the problem itself, which is feasible only at relatively small scales. Second, unlike perceptual data, these problems are often defined on unordered or specially structured inputs. Third, the feasible solutions are subject to a complex set of constraints. These challenges call for dedicated research and the design of differentiable architectures that can (i) well represent the invariances in the input, and (ii) enforce the constraints on the output.
	
	In addition to the scientific challenge of approximating these abstract problems with Machine Learning (ML), solving them has a practical value. The classical combinatorial optimization problems have numerous real world applications, many of which are in management: resource allocation, optimal queuing, task planning, etc. 
	
	The Multiple Traveling Salesmen Problem (mTSP), which is the focus of this work, is a generalization of the well-known Traveling Salesman Problem (TSP). Given a set of cities, $m\geq 1$ salesmen, one depot where salesmen are initially located and to which they return, and a pairwise distance matrix, the objective of the mTSP is to determine a route for each salesman, such that the total length of the routes is minimized, and such that each city is visited exactly once by any of the salesmen. In comparison to the TSP, the mTSP has been the focus of relatively little research. However, as a natural generalization, it captures many more real world problems, ranging from designing satellite surveying to interview scheduling, see~\cite{bektas2006multiple} for a survey.
	
	From the algorithmic perspective, the mTSP poses a significant challenge, in comparison to the TSP. The solution of the TSP is a fixed sequence (up to a circular permutation), which is a well-studied form of output in machine learning. The mTSP requires the computation of a set of sequences of varying lengths, with structural constraints between them. As a result, while the TSP can be solved to some degree of success by methods designed for sequence to sequence generation~\citep{vinyals2015pointer} or even sequence alignment~\citep{levy2017learning}, the mTSP adds an additional layer of complexity.  
	
	Our method encodes the cities, the depot, and the salesmen as three sets and is invariant to the order in each of them. This is done by generalizing the architecture of PointNet~\citep{qi2017pointnet} to a problem involving multiple sets. In addition to this generalization, our method employs what we term Leave-One-Out pooling. To address the spatial layout of the problem, as a planar graph, a learned spatial weighting mechanism is used. On top of the network's layers, a novel scheme is used to ensure that the obtained solution complies with the mTSP, i.e., a novel differentiable subnetwork is added to encourage, in a soft way, the same constraints that an Integer Linear Programming (ILP) formulation of the mTSP enforces on the solution. Combined with a dedicated loss, our method is able to provide better results than any combination of methods in the leading mTSP solver and wins most of the experiments, even if the best baseline method is selected separately, in hindsight, for each experiment.
	
	\section{Related Work}
	\label{sec:prev}
	From the technical perspective, our architecture relates to an emerging body of work, which encodes inputs that are given as sets~\citep{qi2017pointnet,NIPS2017_6931}. Similar to such methods, we employ pooling in order to obtain a representation that is invariant to the order of the elements, followed by a local computation at each element. As shown by~\cite{qi2017pointnet} and~\cite{NIPS2017_6931}, under mild conditions, this is the only way to achieve permutation invariance. As mentioned above, our network is adapted to process multiple sets.
	
	In addition, since the mTSP has properties that are common to both set and graph problems, our network is also inspired by the recent advancements in graph-networks, commonly called \textit{graph-convolutions}~\citep{khalil2017learning,kipf2017semi}. These networks operate on graphs, and use the underlying metric between nodes to propagate information in a hierarchical manner.  While~\cite{khalil2017learning} used the distance metric as an explicit input to each local computation and~\cite{kipf2017semi} used it implicitly in the propagation rule, we use a learned transformation from distances to local weights, which integrates the graph properties to our sets-specialized architecture.
	
	\subsection{TSP solvers}
	
	The TSP is one of the most well-known NP-hard problems, and it has been proven that some instances of the problem, such as the symmetric euclidean variant, are NP complete~\citep{papadimitriou1977euclidean}. As a results, there is no known polynomial time algorithm which solves it. 
	Due to its importance in many real life applications, the TSP has attracted a large amount of research.
	A well-known approximation algorithm for the TSP is due to~\cite{christofides1976worst}, which uses calculation of minimum spanning trees and minimum-weight perfect matching to reach an approximation ratio of 1.5. The Concorde solver~\citep{applegate2006concorde}, which uses highly optimized and carefully crafted methods to efficiently prune the search space, is commonly regarded as the current best exact TSP solver. 
	
	The Pointer Network~\citep{vinyals2015pointer}, which is a general sequence to sequence (seq2seq) architecture, has reignited the interest in Neural Network based TSP solvers in the modern literature, and was followed by~\cite{levy2017learning} that matched its performance using LSTMs followed by convolutional layers. \cite{bello2016neural} used the Pointer Network in conjunction with Reinforcement Learning (RL) to learn a TSP heuristic in an unsupervised way, and~\cite{khalil2017learning} used the \texttt{structure2vec}~\citep{DaiDaiSon16} graph representation together with Q-learning to do the same, albeit without actively training on inference samples. Concurrently to our work, \cite{kool2018attention} have tackled the TSP using an attention-based encoder-decoder, which was trained using RL and was shown to be State of the Art (SOTA) on the TSP for learning algorithms. They used a similar method to counter the Capacitated Vehicle Routing Problem and its Split Delivery variant.
	
	While it is possible to define the solution of the mTSP as a concatenation of sequences (one sequence per salesman), such a solution does not directly expose the constraints of the problem. In addition, a seq2seq solution is, by definition, not permutation invariant. As shown by~\cite{vinyals2015order}, seq2seq models are extremely sensitive to the order of both their inputs and their outputs. Our initial attempts with such a technique resulted in failures.
	
	We experienced with attention-based pooling for the mTSP to no avail, and found the simpler weighted max-pooling to be much more effective. The weighted pooling utilizes the graph structure of the problem. In this work, we present a supervised approach that, despite the complex structure of the output, learns to solve the mTSP and outperforms the leading mTSP solver, while remaining competitive for the TSP. \cite{khalil2017learning}, \cite{bello2016neural} and~\cite{kool2018attention} did not provide an immediate alternative for the leading TSP solver (Concorde), and thus the jury is still out on whether RL would outperform existing TSP solvers, and on whether it is the (only) right approach for solving combinatorial problems.
	
	\subsection{mTSP solvers}
	
	The mTSP has been subject to relatively little research, compared to the TSP. Many of the existing mTSP heuristics can be applied to a larger set of problems, Vehicle Routing Problems (VRPs), in which additional constraints are presented, such as capacity of the salesmen, time windows in which the cities should be visited, etc. 
	Google OR-Tools~\citep{ortools}, a highly optimized and sophisticated operations research program, provides a routing module, which is capable of solving VRPs, as well as the mTSP. It uses local search to solve the instances, by first using a rather simple heuristic to reach an initial solution (such as greedily extending the current route), and then using a chosen meta-heuristic to navigate through the search space and reach a better solution, such as Greedy Descent, Simulated Annealing and Tabu Search. During the search, OR-Tools is able to use VRP-specific heuristics, such as Lin-Kernighan~\citep{lin1973effective} and 2-opt.
	
	The classical neural network literature on the mTSP, mostly generalizes TSP solutions of the era in which it was written. \cite{wacholder1989neural} reduced the mTSP to the TSP by creating as many depot copies as the number of salesmen. The TSP route is then divided into salesmen, by cutting the route every time a copy of the depot is visited. The solution is recovered from an adjacency matrix of a size, which is the square of the number of non-depot cities and the number of salesmen, generalizing the TSP solution of~\cite{Hopfield1985}. \cite{vakhutinsky1994solving} have addressed the VRP using Elastic Nets and \cite{torki1997competitive} have tackled it using Self Organizing Feature Maps (SOFMs). Both approaches are per-sample optimization methods (i.e., no training is performed), which use a neural network to depict the dynamics of the model, and use a problem-tailored energy function to converge to a local minimum. SOFMs were used later on by~\cite{somhom1999competition} and~\cite{modares1999self} to tackle the minmax variant of the mTSP, where the goal is to find routes for the salesmen, such that the maximal tour length among them is minimal.
	
	Concurrent to our work, both~\cite{nazari2018deep} and~\cite{kool2018attention} have tackled two more general variants of the mTSP: the Capacitated Vehicle Routing Problem, in which a maximum capacity is enforced over each vehicle and each city has a demand, and its Split Delivery variant, in which a vehicle may arrive multiple times at each city. Both approaches used an attention-based encoder-decoder architecture, which was trained using RL and then evaluated on synthetic samples. Their approach outputs the solution as a concatenation of sequences in an auto-regressive manner, as opposed to our sparse output.
	
	\section{Problem Formulation}
	
	In the TSP, there are $n$ cities and a distance metric between them $d:[n]\times [n]\rightarrow\mathbb{R}$. The goal is to find a permutation $\pi$ of the cities, such that the distance covered by a salesman traveling according to $\pi$ is minimal: $d(\pi(n),\pi(1))+\sum_{i=1}^{n-1} d(\pi(i),\pi(i+1))$.
	
	There are a number of ways to generalize the TSP to the multiple traveling salesmen setting. In this work we refer to the single depot variant. In this variant, there are $n$ cities, $m$ salesmen and a distance metric $d : [n] \times [n] \rightarrow \mathbb{R}$ between the cities. All salesmen start at city $1$ (the depot), take a route such that each city, other than the depot, is visited exactly once by any of the salesmen, and all salesmen return to the depot at the end of their tour. We define $\delta_{i,j,k}$ to be 1, if and only if salesman $k$ travels from city $i$ to city $j$, and 0 otherwise. The goal is to minimize the sum of the traveled distances:
		\begin{equation}
		\label{eq:mtsp-cost}
		\sum_{i=1}^{n} \sum_{j=1}^{n} \sum_{k=1}^{m} \delta_{i,j,k}d(i,j)
		\end{equation}
	Subject to the following set of redundant constraints:
	\begin{subequations}
		\label{eq:mtsp-constraints}
			\begin{align}
			\label{eq:mtsp-constraints:from-depot}
			\forall k\in[m] : & \sum_{j=2}^n \delta_{1,j,k}=1\\
			\label{eq:mtsp-constraints:to-depot}
			\forall k\in[m] : & \sum_{i=2}^n \delta_{i,1,k}=1\\
			\label{eq:mtsp-constraints:from-others}
			\forall 2\leq i\leq n : & \sum_{j=1}^n \sum_{k=1}^m \delta_{i,j,k}=1\\
			\label{eq:mtsp-constraints:to-others}
			\forall 2\leq j\leq n : & \sum_{i=1}^n \sum_{k=1}^m \delta_{i,j,k}=1\\
			\label{eq:mtsp-constraints:same-salesman}
			\forall 2\leq r\leq n, k\in[m] : & \sum_{i=1}^n \delta_{i,r,k} = \sum_{j=1}^n \delta_{r,j,k}\\
			\label{eq:mtsp-constraints:subtour-elimination}
			\forall 2\leq i \neq j \leq n : & u_i - u_j + (n - m)\cdot \sum_{k=1}^m \delta_{i,j,k} \leq n - m - 1
			\end{align}
	\end{subequations}%
	These constraints denote that: (\ref{eq:mtsp-constraints:from-depot}) every salesman leaves the depot exactly once; (\ref{eq:mtsp-constraints:to-depot}) every salesman returns to the depot exactly once; (\ref{eq:mtsp-constraints:from-others}) every non-depot city is left exactly once; (\ref{eq:mtsp-constraints:to-others}) all salesmen combined, return only once to each non-depot city; (\ref{eq:mtsp-constraints:same-salesman}) the number of times a salesman visits a non-depot city equals the number of times it leaves the city; and (\ref{eq:mtsp-constraints:subtour-elimination}) no subtours exist (degenerate routes that do not include the depot), using $n-1$ latent variables $u_2,...,u_n$.
	
	In this work, we counter the 2-dimensional Euclidean variant of the mTSP. Since the Euclidean TSP is NP-complete~\citep{papadimitriou1977euclidean}, the variant we tackle is NP-hard, and no efficient algorithm is currently known for solving it. For small problems, Eq.~\ref{eq:mtsp-cost} and~\ref{eq:mtsp-constraints} can be solved using ILP solvers.

	\section{Method}
	\label{sec:method}
	
	We start by describing the blocks we use, in order to define a network over multiple groups, in a way that is invariant to the order of elements within the groups; we extend the basic definition by describing how to incorporate a distance metric between elements into the network. We then describe the network version used for the mTSP. Lastly, we describe the subnetwork that is added on top in order to ensure, in a soft way, the constraints of Eq.~\ref{eq:mtsp-constraints}, and the loss used.
	
	\subsection{Permutation Invariant Pooling Network}
	
	We provide a method for working with $k$ different groups of elements, such that each groups $i$ is of variable length $l_i$:
		\begin{equation}
		G_i=\{x_{i,1},...,x_{i,l_i}\}
		\end{equation}
	We would like our network to be invariant to the order of elements in each group and support groups of variable length. We generalize~\cite{qi2017pointnet} to the case of multiple groups and maintain permutation invariance within each group, when creating a context vector, by using a permutation invariant pooling over each group (such as Max Pooling or Average Pooling).
	
\begin{figure}
		\centering
		\includegraphics[width=1.\linewidth]{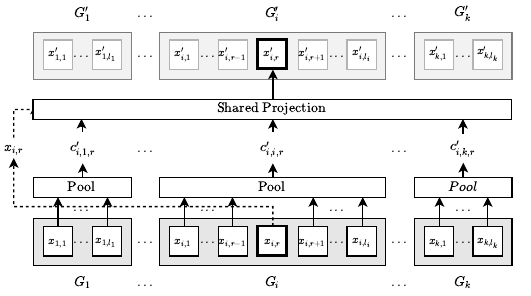}
		\caption{A single permutation invariant pooling layer, without spatial weighting.}
		\label{fig:perm-inv-layer}
	\end{figure}

	In our case, we employ a Leave-One-Out type of pooling. We wish to avoid a situation in which the context vector, which describes a group, overlaps the embedding of a single element in that group. For example, when employing Max Pooling over a group, the vector with the highest value in some coordinate, could benefit from the extra information of receiving the second highest value and not just a copy of the maximal value.
	Therefore, we eliminate \textit{self pooling} by calculating the context from the perspective of each member of the group separately. Formally, we denote by $g$ the permutation invariant pooling function and for each $i,j\in[k]$ and $r\in[l_i]$, we calculate the context:
		\begin{equation} \label{eq:context}
		\begin{gathered}
		c'_{i,i,r} = g(x_{i,1},...,x_{i,r-1},x_{i,r+1},...,x_{i,l_i})\\
		\forall j\neq i, c'_{i,j,r}=g(x_{j,1},...,x_{j,l_j})
		\end{gathered}
		\end{equation}
	The main building block of our network is the \textit{permutation invariant pooling layer}, in which we combine each element $x_{i,r}$ with its contexts $c'_{i,1,r},...,c'_{i,k,r}$ using an affine projection $f_i$, which is shared for every element $x_{i,r}$ of $G_i$:
		\begin{equation}
		x'_{i,r} =  f_i(x_{i,r}||c'_{i,1,r}||...||c'_{i,k,r})
		\end{equation}
	Where $||$ denotes the concatenation operator. The permutation invariant pooling layer is depicted in Fig.~\ref{fig:perm-inv-layer}.
	
	For many problems, there is a given distance metric between some of the elements. For example, in the mTSP, the distance between the cities is given. 
	Inspired by graph-networks, we provide a spatial propagation scheme that is designed for pooling networks:
	Denote by $d_{a,b}$ the distance between elements $a$ and $b$ and by $w(\cdot)$ a learned weighting function. When using weighted Leave-One-Out pooling, we generalize Eq.~\ref{eq:context} and compute the context vector, using the multiplication of the pooled vectors and their corresponding weights:
		\begin{equation}
		\begin{gathered}
		y_{i,j,r,q}=x_{j,q}\odot w(d_{x_{j,q},x_{i,r}})\\
		c'_{i,i,r}=g(y_{i,i,r,1},...,y_{i,i,r,r-1},y_{i,i,r,r+1},...,y_{i,i,r,l_i})\\
		\forall j\neq i, c'_{i,j,r}=g(y_{i,j,r,1},...,y_{i,j,r,l_j})
		\end{gathered}
		\end{equation}
	Where $\odot$ denotes element-wise multiplication.
	
	Following \cite{NIPS2017_7181}, each permutation invariant pooling layer is followed by a fully connected network composite of a single hidden layer, followed by the ReLU activation function. We set the hidden layer size to be $d_{ff}$. The fully connected network is shared across its corresponding group, similar to the sharing of the projection $f_i$, and thus we actually have $k$ different fully connected networks. The permutation invariant pooling layer and the shared fully connected network are each wrapped in a residual block and are followed by a layer normalization:
		\begin{equation}
		x'=Norm(h(x)+x)
		\end{equation}
	Where $h$ is either the permutation invariant pooling layer or the shared fully connected layer, and $Norm$ is the layer normalization operator~\citep{ba2016layer}, 
	which removes from each activation the mean and divides by the standard deviation of all the activations in the layer, for the sample being evaluated.
	
\begin{figure}
		\centering
		\includegraphics[scale=2.5]{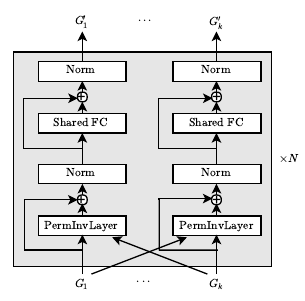}
		\caption{The generic network architecture.}
		\label{fig:perm-inv-net}
	\end{figure}

	A single block, which is composed of the permutation invariant pooling layer and the shared fully connected layer, is depicted in Fig.~\ref{fig:perm-inv-net}.
	
	\subsection{Learning the Multiple Traveling Salesmen Problem}
	
	Utilizing the permutation invariant pooling network for the mTSP is done by representing each instance of the problem as three groups: a group of salesmen, a singleton which contains the depot and a group of the other cities (note that unlike the TSP, in the mTSP, the depot and the other cities play a slightly different role). Denote by $USV^\top$ the $d_{svd}$-dimensional approximation of the inter-city distance matrix. We encode the cities as the rows of $US$, and encode each salesman $k\in[m]$ using the 2-dimensional vector $(k/m, m)$, embedding both local and global information about the salesman.
	
	The calculation of the network starts with a shared affine embedding of each input element into a vector of size $d_{model}$, where the sharing is between elements of the same group, followed by a layer normalization. The groups are then fed into $N$ consecutive permutation invariant pooling blocks, to produce a complex hidden representation of each salesman and city (including the depot). The size of the representations of the salesmen and the groups remains $d_{model}$ throughout the $N$ blocks.
	
	Finally, for each $i,j\in [n]$ and $k\in [m]$ we concatenate the representation of the $i$-th city, $j$-th city and $k$-th salesman to get a hidden representation of size $3d_{model}$ of the path from the $i$-th city to the $j$-th city by the $k$-th salesman. We employ a shared fully-connected network over each of these representations to get a multi-layer adjacency tensor of size $m\times n\times n$, i.e., we employ a fully connected network with a single hidden layer of size $d_{model}$, followed by a ReLU and a scalar output to obtain a score for a salesman $k$ making the path from city $i$ to city $j$.
	
	\subsubsection{Adhering to the mTSP Constraints}
	\label{sec:beam-search}
	
	The output of the network is an $m \times n \times n$ tensor of real values. In order to make the output more interpretable and to make the entire model differentiable, we relax the restriction that the output of the network must be binary and allow it to be a real value between 0 and 1. By applying a multi-dimensional generalization of the common softmax-based normalization scheme, in a way that generalizes the Softassign method by~\cite{gold1996softassign}, we next transform the network's output into an estimation of what we call a \textit{semi multi-stochastic tensor}. 
	
	The semi multi-stochastic tensor is a tensor of real values between 0 and 1, such that Eq. \ref{eq:mtsp-constraints:from-depot} to \ref{eq:mtsp-constraints:to-others} hold. We obtain this tensor via Alg.~\ref{alg:softassign}. In each iteration, the algorithm satisfies two constraints: In the odd iterations lines 5--10 are run, and Eq.~\ref{eq:mtsp-constraints:from-depot} and~\ref{eq:mtsp-constraints:from-others} are fulfilled, and in the even iterations, the section of lines 12--17 runs and fulfills Eq.~\ref{eq:mtsp-constraints:to-depot} and~\ref{eq:mtsp-constraints:to-others}. For example, in line 6 each $x^{r-1}_{k,1,j}$ is being scaled by a constant, such that Eq.~\ref{eq:mtsp-constraints:from-depot} holds. Note that in each iteration, at least two equations hold, while possibly creating a violation of the other two constraints. We believe that by generalizing the proof of~\cite{sinkhorn1967concerning}, one could prove our empirical observation that the method converges to a tensor that simultaneously satisfies all four constraints.

\begin{algorithm}[t]
	\caption{Softassign for the mTSP}
	\label{alg:softassign}
		\begin{algorithmic}[1]
			\STATE {\bfseries Input:} tensor $o_{k,i,j}\in \mathbb{R}^{m\times n\times n}$, \#iterations $T$
			\STATE Initialize $(x^0_{k,i,j})=(exp(o_{k,i,j}))$. \COMMENT{Ensure positivity}
			\FOR{$r=1$ {\bfseries to} $T$}
			\IF{$r\ \mathrm{mod}\ 2 = 1$}
			\STATE \COMMENT{Satisfy Eq. \ref{eq:mtsp-constraints:from-depot}}
			\STATE Set $(x^r_{k,1,j})=(x^{r-1}_{k,1,j}/\sum_{j'=1}^n x^{r-1}_{k,1,j'})$. 
			\FOR{$i=2$ {\bfseries to} $n$}
			\STATE \COMMENT{Satisfy Eq. \ref{eq:mtsp-constraints:from-others}}
			\STATE Set $(x^r_{k,i,j})=(x^{r-1}_{k,i,j}/\sum_{k'=1}^m \sum_{j'=1}^n x^{r-1}_{k',i',j})$. 
			\ENDFOR
			\ELSE
			\STATE \COMMENT{Satisfy Eq. \ref{eq:mtsp-constraints:to-depot}}
			\STATE Set $(x^r_{k,i,1})=(x^{r-1}_{k,i,1}/\sum_{i'=1}^n x^{r-1}_{k,i',1})$. 
			\FOR{$j=2$ {\bfseries to} $n$}
			\STATE \COMMENT{Satisfy Eq. \ref{eq:mtsp-constraints:to-others}}
			\STATE Set $(x^r_{k,i,j})=(x^{r-1}_{k,i,j}/\sum_{k'=1}^m \sum_{i'=1}^n x^{r-1}_{k',i',j})$. 
			\ENDFOR
			\ENDIF
			\ENDFOR
			\STATE {\bfseries Return} $(x^T_{k,i,j})$.
		\end{algorithmic}
	\end{algorithm}

	After running Alg.~\ref{alg:softassign} for $T$ iterations, a \textit{soft} multi-layer adjacency tensor is obtained. In order to find the routes with the highest probability, with respect to the net's output, a search is performed. Specifically, we use a beam search to find the $b$ most probable valid solutions, and from them choose the best one as the final solution. The details of the beam search we use are as follows, where $z_{k,i,j}$ is the output of Alg.~\ref{alg:softassign}:
	
	First, the output corresponding to the paths exiting the depot are considered ($z_{k,1,j}$), and a separate beam search is being employed in order to find the top $b$ options for the salesmen's starts; each option includes for each salesman the city to which it will go after it leaves the depot. Formally, the first-phase beam search finds the top $b$ options $S_1,...,S_b$ maximizing the following probability-like value:
	\begin{equation}
	\text{START-VALUE}(S_i)=\prod_{k=1}^{m}z_{k,1,S_{i,k}}
	\end{equation}
	After the first pool of such $b$ incomplete solutions has been acquired, the following process is executed: For each incomplete solution in the current pool, denote by $k$ the ordinal number of the first salesman whose route has not yet ended (its final city is currently not the depot). There are at most $n$ options for the next city in such salesman's route, which do not violate Eq.~\ref{eq:mtsp-constraints}, i.e., such that every city is not visited more than once, etc.; for every such possible option we add a new, possibly incomplete solution to the new solutions pool, which is composed of the current solution with the route of salesman $k$ continuing to the next chosen city. Applying the same procedure to all $O(b)$ solutions in the current pool results in a new pool of $O(b\cdot n)$ solutions. These solutions are pruned so that only the top $b$ solutions with the highest values (the current value multiplied by the cell in $z$ corresponding to the new path added) are left. We repeat this process until $O(b)$ full, valid solutions are found. The final solution with the shortest sum of traveled distances constitutes the final output. Note that in order to make the beam search more numerically-stable, we maximize the sum of the log of the probabilities instead of their product.
	
	\subsubsection{Representation Invariant Loss}
	\label{sec:loss}
	
	Since we want both the input and the output of the network to be invariant to the representation of the problem, i.e. the order of the cities, the order of the salesmen and the directions in which they travel, we use a custom loss function, inspired by~\cite{vinyals2015order}. We modify the target of each sample of $n$ cities and $m$ salesmen and obtain $2^mm!$ possible targets, which consist of every possible order of the salesmen and every possible routes' directions. Every target is encoded as a multi-layer adjacency tensor of size $m\times n\times n$. The loss of a single sample of $n$ cities and $m$ salesmen is then defined as the minimum normalized negative log-likelihood of the output of Alg.~\ref{alg:softassign}, denoted by $z_{k,i,j}$, with respect to the possible targets:
		\begin{equation}
		\label{eq:native-loss}
		\begin{gathered}
		\begin{aligned}
		\mathcal{L}=\min_{\textbf{b}\in {\{0,1\}}^m,\pi} -\Big(\frac{1-\lambda}{n-1}\sum_{i=2}^n \sum_{j=1}^n \sum_{k=1}^m \log z_{k,i,j} \cdot t^{b_k}_{\pi(k),i,j}+\frac{\lambda}{m}\sum_{j=1}^n \sum_{k=1}^m \log z_{k,1,j} \cdot t^{b_k}_{\pi(k),1,j}\Big)
		\end{aligned}\\
		t^0_{k,i,j}=t_{k,i,j}, t^1_{k,i,j}=t_{k,j,i}
		\end{gathered}
		\end{equation}
	where $\lambda\in [0,1]$ is the relative weight given to the paths exiting the depot and $t_{k,i,j}=1$ if salesman $k$ travels from city $i$ to city $j$ in the target label and 0 otherwise.

	The implementation of the na\"ive definition of our representation invariant loss, as defined in Eq.~\ref{eq:native-loss}, requires $O(2^m\cdot m!\cdot m)$ matrix multiplications of size $n\times n$ (by multiplying each of the $m$ layers of the output by their corresponding target layers in every possible representation of the target, after applying negative log). The optimal routes directions are easy to compute once the permutation of the salesmen is set, and thus we can pre-calculate the optimal loss between each pair of output layer $k$ and target layer $p$:
	\begin{equation}
	\label{eq:loss-pre}
	L_{k,p}=\min_{b\in\{0,1\}} -\Big(\frac{1-\lambda}{n-1}\sum_{i=2}^n \sum_{j=1}^n \log z_{k,i,j} \cdot t^{b_k}_{p,i,j} + \frac{\lambda}{m}\sum_{j=1}^n \log z_{k,1,j} \cdot t^{b_k}_{p,1,j}\Big)
	\end{equation}
	And get the following, equivalent definition of the loss:
	\begin{equation}
	\label{eq:loss-post}
	\mathcal{L}=\min_{\pi} \sum_{k=1}^m L_{k,\pi(k)}
	\end{equation}
	Calculating Eq.~\ref{eq:loss-pre} requires $O(m^2)$ matrix multiplications of size $n\times n$, while computing Eq.~\ref{eq:loss-post} does not require any matrix multiplications but needs $O(m!\cdot m)=O((m+1)!)$ simple operations. For a constant $m$ this makes the loss calculation polynomial. Even when $m$ is a variable, the loss is calculated only during training, hence the algorithm implemented by the network is polynomial during inference.

	The entire model, including Alg.~\ref{alg:softassign}, is differentiable, and, therefore, the model is trained in an end-to-end manner using samples of variable sizes, which have been solved to optimality by an ILP solver.
	
	\section{Experiments}
	\label{sec:experiments}

	\subsection{Datasets}
	To enable supervised training, we generate $\sim7.3$ million random mTSP instances in the following way: for each $m$ between $1$ and $5$, for each $n$ between $\min(4,2m)$ and $20$, we uniformly sample $n$ random cities in the unit square, set the first city sampled to be the depot and $m$ to be the number of salesmen. We use the ILP formulation described in Eq.~\ref{eq:mtsp-cost} and~\ref{eq:mtsp-constraints} in conjunction with Matlab's built-in \textit{intlinprog} in order to solve each instance to optimality. We generate 99,000 instances for each such combination of $n$ and $m$ to be used as a training set (\textbf{\textit{mTSP-train}}), and 1,000 instances to be used as a test set (\textbf{\textit{mTSP-test}}). Out of the 99,000 samples in the training set, we use 1,000 as the validation set for hyper-parameters cross validation.
	
	In addition to \textit{mTSP-train} and \textit{mTSP-test}, we use the \textbf{\textit{mTSPLib}} benchmark defined by~\cite{necula2015tackling}. This benchmark reuses samples from the public dataset \textit{TSPLib}~\citep{reinelt1991tsplib}, which contains various real-life instances of the TSP. Specifically, \textit{mTSPLib} is created from \textit{TSPLib}, by taking four problems (\textit{eil51}, \textit{berlin52}, \textit{eil76} and \textit{rat99}, which contain 51, 52, 76 and 99 cities respectively), and from each of them defining four mTSP samples by setting the first city in the cities' list to be the depot and the number of salesmen to be 2, 3, 5 or 7. Therefore, \textit{mTSPLib} consists of 16 individual mTSP samples, whose size is much larger than those of \textit{mTSP-train}. Created this way, our network has to generalize in two ways in order to be competitive: (1) work with considerably bigger problems, and (2) be effective for problems created by a completely different process.
	
	In addition, we test our network on three TSP benchmarks first presented in \cite{vinyals2015pointer}: \textbf{\textit{TSP5}}, \textbf{\textit{TSP10}} and \textbf{\textit{TSP20}}, where \textit{TSP}X contains 10,000 TSP instances of \textit{X} cities. We compare our results on these benchmarks to the public results of \cite{vinyals2015pointer,levy2017learning,bello2016neural,kool2018attention}.
	
	\subsubsection{Operations Research Baselines}
	We compare ourselves with the routing module of OR-Tools~\citep{ortools}, which supports a large number of configurations and heuristics designed for routing problems, such as the VRP and the mTSP, and is commonly used to solve real-life instances of such problems. OR-Tools uses local search in order to solve routing problems: it starts by finding an initial solution for the problem (which may be suboptimal), and continues with a predefined metaheuristic to locally navigate through the solution space, trying to improve its current best solution.

	The different strategies provided in OR-Tools, which may be used to find the first solution, are diverse and contain both simple rules-of-thumb and more sophisticated heuristics, including: (1)~{\textbf{\textit{path--cheapest--arc}}}, which iteratively extends the current route by choosing the closest city as the next city; (2)~{\textbf{\textit{path--most--constrained--arc}}}, which iteratively extends the current route using a comparison-based selector that favors the most constrained "arc" (path between two cities); (3)~\textbf{{\textit{global--cheapest--arc}}}, which iteratively connects two previously disconnected cities whose distance is minimal; (4)~\textbf{{\textit{local--cheapest--arc}}}, which iteratively connects the first city in-order that has no successor to a free city, such that the distance between them is minimal; and (5)~\textbf{{\textit{first--unbound--min--value}}}, which connects the first city without a successor to the first free city. The other strategies provided in OR-Tools, which are more specialized to the TSP, did not work when solving mTSP instances.

	The local-search metaheuristics used in OR-Tools are as follows: (1)~\textbf{\textit{Greedy Descent}}, which accepts neighbor solutions only if the solution's cost decreases, until a local minima is reached; (2)~\textbf{\textit{Guided Local Search}}, which uses a penalized cost function in order to escape local minimas and plateaus; (3)~\textbf{\textit{Simulated Annealing}}, which uses decreasing temperatures in order to balance the exploration and exploitation during the search; (4)~\textbf{\textit{Tabu Search}}, which uses a restriction mechanism in order to escape local minimas; and (5)~\textbf{\textit{Objective Tabu Search}}, which uses Tabu Search on the cost of the solution, instead of the solution itself.

	In each testing scenario, we compare ourselves to two OR-Tools based models: OR@25 and {\sc ORmin}. OR@25 is the ensemble, which consists of all 25 possible combinations of a first strategy and a meta-heuristic for a given solutions limit, such that the final solution chosen is the best one out of the 25 (possibly) different solutions found. {\sc ORmin} is the combination of a first strategy and a meta-heuristic that out of the 25 possible combinations had the best performance on the dataset considered. I.e, {\sc OR@25} is chosen per sample, while {\sc ORmin} is chosen per experiment. By definition, OR@25 is always better than {\sc ORmin}, and {\sc ORmin} may change between different experiments.
	
	\subsubsection{The Neural Network Baseline}
	As explained in Sec.~\ref{sec:prev}, the Pointer Network has no straight-forward generalization to the mTSP, in a way that incorporates the problem constraints. In order to obtain a neural baseline, we generalize the work of~\cite{levy2017learning} to the multiple salesmen setting, by introducing a new salesmen dimension to the two existing dimensions of source cities and destination cities.
	The encodings in the three dimensions are passed through LSTMs and then are concatenated, in a three dimensional grid, to produce a four dimensional encoding of each sample (the fourth dimension being the number of channels in the encoding).
	We pass this tensor through a three dimensional CNN, to get the final multi-layer output, and then continue using the same pipeline as our own (Alg.~\ref{alg:softassign} and then our custom loss). We use the same hyper-parameters reported in~\cite{levy2017learning}, and encode the cities using their Cartesian coordinates. For this baseline network, positional encoding~\citep{NIPS2017_7181} outperformed our own as the salesmen encoding, and it is the one being reported.
	
	\subsubsection{Our Network}
	
	Our network is described in Sec.~\ref{sec:method}. We set $d_{svd}=4$, $d_{model}=256$, $d_{ff}=1024$ and $N=7$, and the weight inside the loss to be $\lambda=0.5$. We empirically set the number of iterations of Alg.~\ref{alg:softassign} to be $T=100$. For the permutation invariant pooling $g$, we choose Max Pooling, and set $w(d)=A\odot exp(-C\cdot d)+B$, where $A$, $B$ and $C$ are learned vectors of size $d_{model}$. We use weighted pooling layers, whenever we pool from a group of cities (or the depot) over other cities; otherwise, when salesmen are involved, we use regular pooling layers. In order to maintain scale invariance, we divide the distance matrix by its mean before applying the SVD approximation. In the experiments on the \textit{mTSP-test} benchmark, we use a beam search, as described in Sec.~\ref{sec:beam-search}. However, for \textit{mTSPLib}, in order to improve performance and generalizability, we test our network as a first strategy for the OR-Tools pipeline; to assure fairness, and because our network uses a beam search as its last decoding stage, when combined with an OR-Tools meta-heuristic, we define the pipeline's beam size to be the sum of the solutions checked by our beam search and the ones checked by the local search method. We found that allocating 10\% of the beam size to the beam search and the rest to the local search works well, where the best meta-heuristic as a follow up is Guided Local Search. This configuration defines \textit{Our Pipeline} when experimenting with \textit{mTSPLib}.
	
	\subsubsection{Ablation Study}
	
	\begin{table*}[b]
		\caption{Average error on \textit{mTSP-test}, measured as the ratio between the obtained sum of the routes' lengths and the optimal one, minus one.}
		\label{table:mtsp-test}
		\centering
		\begin{small}
			\begin{sc}
				\begin{tabular}{lcccc}
					\toprule
					Model & Beam 1 & Beam 20 & Beam 200 & Beam 2,000 \\
					\midrule
					NN Baseline & 28.44\% & 15.12\% & 9.24\% & 5.97\% \\
					ORmin & 21.76\% & 2.51\% & 0.14\% & \textbf{$<$0.01\%} \\
					\midrule
					Our w/o invariant loss & 6.59\% & 0.57\% & 0.02\% & \textbf{$<$0.01\%} \\
					Our w/o spatial weighting & 1.79\% & 0.07\% & 0.01\% & \textbf{$<$0.01\%} \\
					Our w/o LOO in the pooling& 1.85\% & 0.04\% & \textbf{$<$0.01\%} & \textbf{$<$0.01\%} \\
					Our & \textbf{0.95\%} & \textbf{0.01\%} & \textbf{$<$0.01\%} & \textbf{$<$0.01\%} \\
					\bottomrule
				\end{tabular}
			\end{sc}
		\end{small}
	\end{table*}
	
		\begin{table*}
        \caption{Number of samples of \textit{mTSPLib} in which the method is better than OR@25 (out of 16).}
        \label{table:mtsplib-ablation}
        \centering
        \begin{small}
        \begin{sc}
        \begin{tabular}{@{}lc@{~~~}c@{~~~}c@{~~~}c@{}}
        \toprule
        Beam Size & 1 & 20 & 200 & 2,000 \\
        \midrule
        Our w/o spatial weighting & 11 & 10 & 0 & 1 (+3 equal) \\
        Our w/o LOO in pooling & 8 & 7 & 4 & 2 (+4 equal) \\
        \midrule
        Our & \textbf{12} & \textbf{12} & \textbf{7} & \textbf{4} (+4 equal) \\
        \bottomrule
        \end{tabular}
        \end{sc}
        \end{small}
    \end{table*}
    
    \begin{table*}
		\caption{Comparison of route lengths for various methods on the instances of \textit{mTSPLib}. $m$ is the number of salesmen. {\sc ORmin} starts for all beam sizes with \textit{path--cheapest--arc} and follows it by Guided Local Search.}
		\label{table:mtsplib}
		\begin{ssmall}
			\begin{sc}
				\begin{tabular}{@{}l@{~}l@{~~}c@{~}c@{~}c@{~~}c@{~}c@{~}c@{~~}c@{~}c@{~}c@{~~}c@{~}c@{~}c@{}}
					\toprule
					& & \multicolumn{3}{c}{Beam 1} & \multicolumn{3}{c}{Beam 20} & \multicolumn{3}{c}{Beam 200} & \multicolumn{3}{c}{Beam 2,000} \\
					\cmidrule(lr){3-5} 
					\cmidrule(lr){6-8} 
					\cmidrule(lr){9-11} 
					\cmidrule(lr){12-14} 
					Data & $m$ & OR@25 & ORmin & Our & OR@25 & ORmin & Our & OR@25 & ORmin & Our & OR@25 & ORmin & Our\\
					\midrule
					\multirow{4}{*}{eil51} 
					& 2 & \textbf{517.20} & 581.18 & 552.55 & 459.81 & 469.13 & \textbf{458.63} & \textbf{436.10} & 438.49 & 446.71 & \textbf{435.18} & 436.80 & \textbf{435.18} \\
					& 3 & 531.15 & 622.30 & \textbf{516.34} & 468.62 & 509.96 & \textbf{458.62} & \textbf{447.28} & 450.73 & 447.99 & \textbf{445.99} & \textbf{445.99} & \textbf{445.99} \\
					& 5 & 624.57 & 660.96 & \textbf{562.01} & 515.12 & 515.12 & \textbf{504.60} & \textbf{472.24} & 476.94 & 480.90 & \textbf{471.69} & \textbf{471.69} & 472.24 \\
					& 7 & 682.72 & 752.92 & \textbf{562.02} & 595.02 & 644.08 & \textbf{552.22} & \textbf{512.60} & \textbf{512.60} & 512.98 & \textbf{508.70} & \textbf{508.70} & \textbf{508.70} \\
					\midrule
					\multirow{4}{*}{berlin52}
					& 2 & 10113.98 & 10242.83 & \textbf{9144.95} & 8961.63 & 8961.63 & \textbf{8493.03} & 7902.75 & 7922.77 & \textbf{7796.42} & \textbf{7632.43} & 7632.73 & 7649.25 \\
					& 3 & 10879.54 & 11373.27 & \textbf{9607.96} & 9062.35 & 9062.35 & \textbf{8818.54} & 7923.76 & 7989.57 & \textbf{7920.41} & \textbf{7737.02} & 7737.31 & 7737.31 \\
					& 5 & 11537.96 & 12423.95 & \textbf{10318.19} & 10998.44 & 10998.44 & \textbf{9509.13} & \textbf{8125.98} & 8224.63 & 8270.22 & \textbf{8125.98} & \textbf{8125.98} & 8126.28 \\
					& 7 & 13064.96 & 13336.15 & \textbf{9736.35} & 12139.20 & 12749.66 & \textbf{9134.81} & \textbf{8585.41} & 8710.06 & 8641.24 & \textbf{8585.41} & \textbf{8585.41} & \textbf{8585.41} \\
					\midrule
					\multirow{4}{*}{eli76}
					& 2 & 685.02 & 689.19 & \textbf{630.87} & \textbf{615.06} & \textbf{615.06} & 687.40 & 561.90 & 561.90 & \textbf{555.72} & 553.83 & 555.44 & \textbf{552.46} \\
					& 3 & \textbf{723.59} & \textbf{723.59} & 727.53 & \textbf{645.33} & \textbf{645.33} & 662.38 & 568.37 & 568.37 & \textbf{563.24} & \textbf{561.11} & 564.07 & 562.11 \\
					& 5 & 757.13 & 757.13 & \textbf{698.80} & 655.76 & 655.76 & \textbf{614.79} & \textbf{590.04} & 591.64 & 595.27 & 581.43 & 587.34 & \textbf{581.35} \\
					& 7 & 789.38 & 789.38 & \textbf{722.79} & 688.35 & 688.35 & \textbf{688.25} & 623.60 & 623.60 & \textbf{619.49} & 618.28 & 619.41 & \textbf{612.66} \\
					\midrule
					\multirow{4}{*}{rat99} 
					& 2 & \textbf{1638.61} & \textbf{1638.61} & 1909.92 & \textbf{1546.46} & \textbf{1546.46} & 1876.35 & 1271.30 & 1311.54 & \textbf{1265.51} & \textbf{1247.89} & 1257.45 & 1257.32 \\
					& 3 & \textbf{1854.05} & \textbf{1854.05} & 1995.64 & \textbf{1765.51} & \textbf{1765.51} & 1921.73 & \textbf{1298.29} & 1341.05 & 1334.98 & \textbf{1276.97} & 1294.16 & 1277.86 \\
					& 5 & 2397.25 & 2397.25 & \textbf{2081.82} & 2296.52 & 2296.52 & \textbf{2031.07} & 1402.93 & 1405.64 & \textbf{1380.28} & \textbf{1362.58} & 1366.77 & 1378.25 \\
					& 7 & 2861.38 & 2988.40 & \textbf{2094.04} & 2710.29 & 2906.77 & \textbf{1911.32} & \textbf{1491.35} & 1559.66 & 1525.10 & 1471.90 & 1488.53 & \textbf{1471.84} \\
					\bottomrule
				\end{tabular}
			\end{sc}
		\end{ssmall}
	\end{table*}
	
			\begin{table*}
		\caption{Average running time on a sample from \textit{mTSPLib} in seconds.
			{\sc ORmin} starts with \textit{PATH--CHEAPEST--ARC} and follows by Guided Local Search, as defined in Tab.~\ref{table:mtsplib}.}
		\label{table:time}
		\centering
		\begin{small}
			\begin{sc}
				\begin{tabular}{lcccc}
					\toprule
					Beam size & 1 & 20 & 200 & 2,000 \\
					\midrule
					ORmin & 0.16 & 0.17 & 3.66 & 54.04 \\
					Our Pipeline & 0.27 & 0.29 & 4.19 & 54.46 \\
					\bottomrule
				\end{tabular}
			\end{sc}
		\end{small}
	\end{table*}
	
	To show that our method is both sufficient and necessary, we test the necessity of three integral parts of our network: (1) the representation invariant loss, (2) the weighted pooling layers, and (3) the Leave-One-Out pooling. We train three models, each with the corresponding part absent, and test their performance relative to our complete model, on both \textit{mTSPLib} and \textit{mTSP-test}.
	
	\subsubsection{Training}
	
	We train all neural networks on \textit{mTSP-train}, using the Adam learning rate scheme \citep{kingma2014adam}, with the parameters $\beta_1=0.9$, $\beta_2=0.999$ and $\epsilon=1e-8$; The NN baseline is trained on mini batches of 32 samples, each using a constant learning rate of 0.001, while our networks are trained on mini batches of 128 samples, using a constant learning rate of 0.0001. When we are not using the SVD approximation of the cities, we normalize the cities' representations in the dataset, such that their coordinates are uniformly distributed between -1 and 1, using the affine transformation $c'=2c-1$. For the networks which are not invariant to the order of the cities (other than the depot) and the direction of the tours, such as the NN baseline, we randomly permute the order of the cities.
	
	\subsubsection{Results}
	
		\begin{table*}
		\caption{Average tour length on the TSP benchmarks provided by \cite{vinyals2015pointer}. \cite{bello2016neural} does not use a beam search, but an Active Search method that samples 1,280,000 solutions.}
		\label{table:tsp}
		\centering
		\begin{small}
			\begin{sc}
				\begin{tabular}{lccc}
					\toprule
					& \textit{TSP5} & \textit{TSP10} & \textit{TSP20} \\
					\midrule
					Optimal & \textbf{2.12} & \textbf{2.87} & \textbf{3.82}\\
					\midrule
					\cite{vinyals2015pointer} (Beam 20) & \textbf{2.12} & \textbf{2.87} & 3.88 \\
					\cite{levy2017learning} (Beam 10)& \textbf{2.12} & 2.88 & - \\
					\cite{bello2016neural} (Beam 1,280,000) & - & - & \textbf{3.82} \\
					\cite{kool2018attention} (Beam 1,280) & - & - & 3.83 \\
					\midrule
					Our (Beam 1) & \textbf{2.12} & \textbf{2.87} & 3.95 \\
					Our (Beam 20) & \textbf{2.12} & \textbf{2.87} & 3.84 \\
					\bottomrule
				\end{tabular}
			\end{sc}
		\end{small}
	\end{table*}
	
\begin{figure}
        \centerline{\includegraphics[width=1\linewidth]{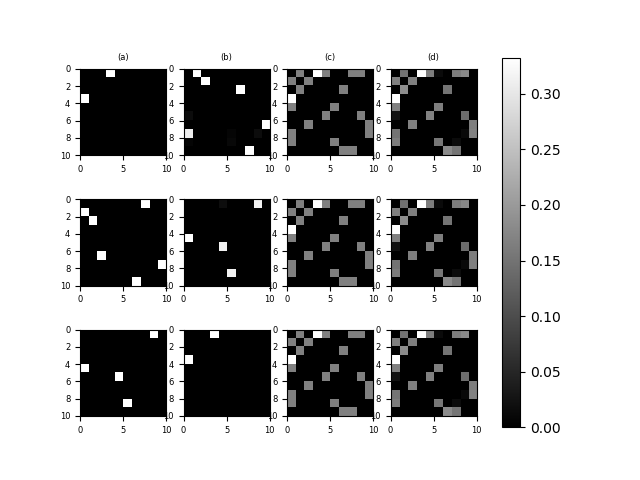}}
        \vskip -0.5in
        \caption{Qualitative comparison of the representation invariant loss, on a sample with $n=10$ and $m=3$ from \textit{mTSP-test}. Each figure is an adjacency matrix, and each column composes a full solution ($m$ adjacency matrices). Permuting a column corresponds to changing the order of the salesmen, and transposing a matrix corresponds to reversing the direction of a salesman's route. Left to right: (a) The optimal solution, as solved by an ILP solver, (b) The output of our network, (c) the mean of the optimal solutions of the different salesmen and their transpositions, (d) the output of our network without the representation invariant loss.}
        \label{fig:perm-inv-loss}
    \end{figure}

	Tab.~\ref{table:mtsp-test} shows our performance on \textit{mTSP-test}. Our method outperforms both the NN baseline and the best OR-Tools method, selected separately for each beam size. Quantitatively, it is seen that the representation invariant loss significantly contributes to the overall success. A qualitative comparison between our network with and without the representation invariant loss is shown in Fig.~\ref{fig:perm-inv-loss}: our network with the custom loss learns correctly to output a valid representation of the optimal solution, where permutation of the rows corresponds to a salesmen permutation and transposition of each matrix corresponds to changing the direction of the route ((a) and (b)). However, when we use a na\"ive negative log likelihood loss, our network seems to predict for each salesmen a route which is the mean of the optimal routes for each salesman and their transpositions ((c) and (d)); this can be explained by the fact that permuting the salesmen and reversing their routes maintains the optimality of the solution, and thus with big enough dataset and inadequate loss function, the network will get stuck on the mean of all possible representations of the optimal solution, which constitutes a local minima.
	
	While one might argue that the experiments on \textit{mTSP-test} show only a small degradation in the performance of our network when either spatial weighting or Leave-One-Out pooling are not used, Tab.~\ref{table:mtsplib-ablation} shows that the performance of those models on \textit{mTSPLib} is much worse. Using spatial weighting and Leave-One-Out pooling helps the generalizability of the network, and both methods are crucial in order to be competitive on \textit{mTSPLib} when using larger beam sizes.

	The raw results on the \textit{mTSPLib} benchmark, which contains bigger and more diverse samples, are shown in Tab.~\ref{table:mtsplib}. For relatively small beam sizes, more often than not, our method outperforms the maximal performance obtained by any of the OR-Tools methods at each individual experiments.

\begin{figure}
        \centerline{\includegraphics[width=0.85\linewidth]{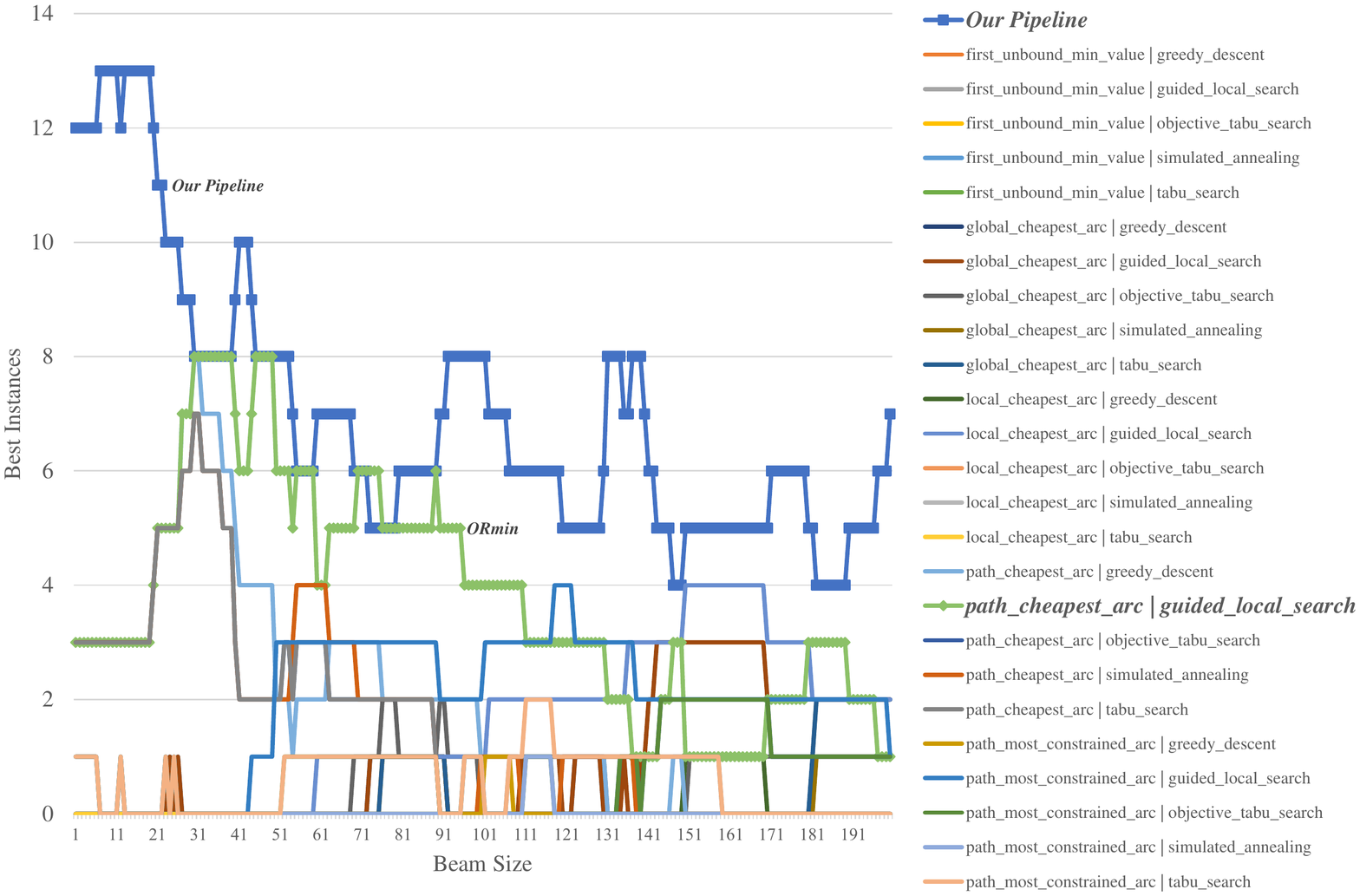}}
        \vskip -0.5in
        \caption{The number of "best" instances, when comparing our pipeline to all OR-Tools combinations on \textit{mTSPLib}, for beam sizes between 1 and 200. An instance is "best" for a method, if no method reaches better results for the instance. Best viewed in color.}
        \label{fig:mtsplib-dominant}
    \end{figure}

\begin{figure}
        \centerline{\includegraphics[width=0.85\linewidth]{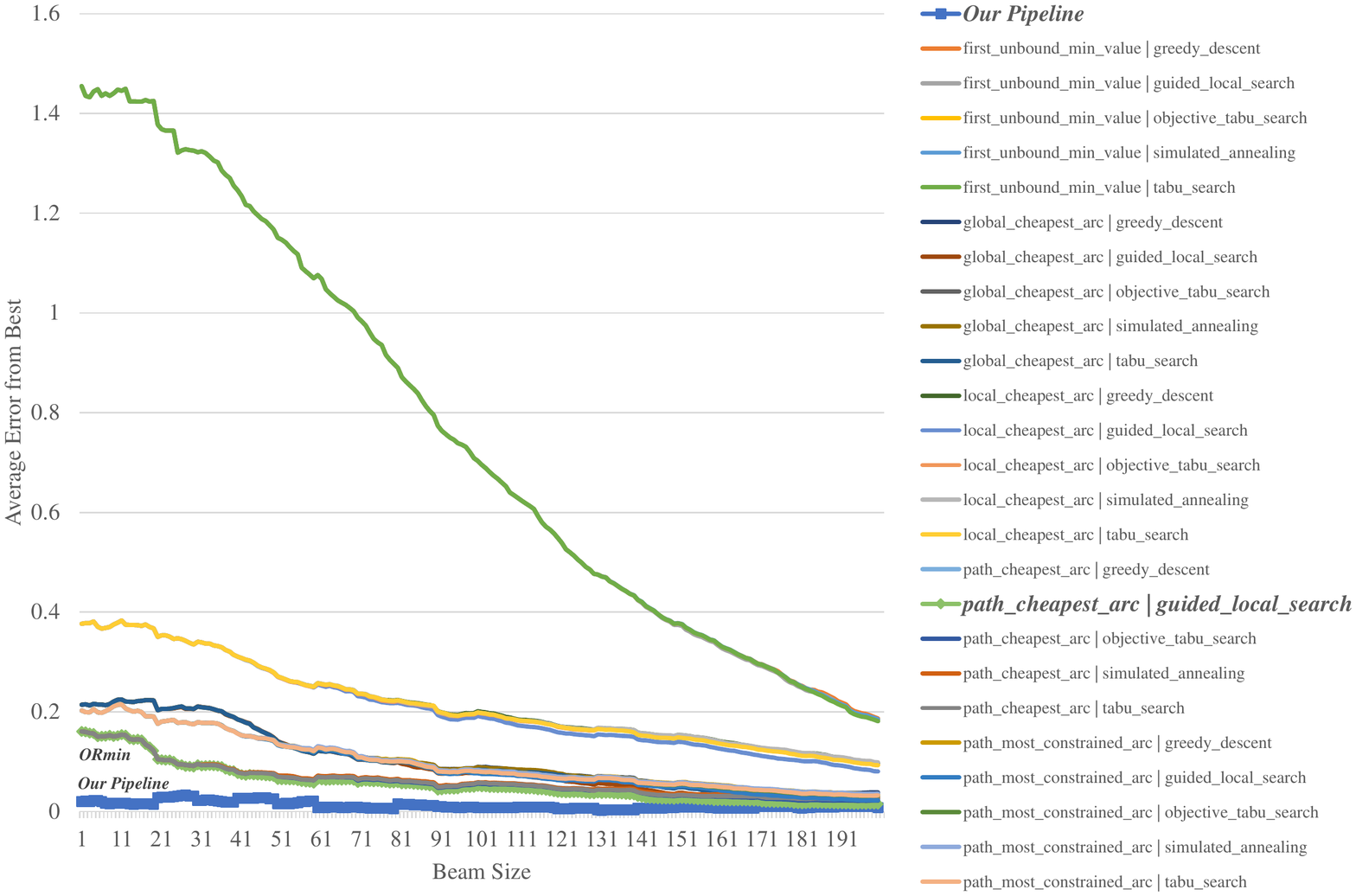}}
        \vskip -0.5in
        \caption{The average error between a method and the solution of the best method for that beam size on \textit{mTSPLib}, out of all OR-Tools method combinations and our pipeline, for beam sizes between 1 and 200. Best viewed in color.}
        \label{fig:mtsplib-ratio}
    \end{figure}

	Fig.~\ref{fig:mtsplib-dominant} presents our comparison of the number of "best instances" of our pipeline and those of the different OR-Tools methods, where an instance is called "best" for a method if for the given beam size, no other method reaches better results. It shows that although the number of instances in which we are strictly better than OR@25 gets smaller as the beam size increases (Tab.~\ref{table:mtsplib}), we remain the best option when regarding a single method only. Fig.~\ref{fig:mtsplib-ratio} shows the average error between each method (either ours or OR-Tools') and the best solution achieved by a method for that beam size. For smaller beam sizes our pipeline has marginally lower average error, and as the beam size grows the difference gets smaller, but we remain the best alternative over the other methods of OR-Tools.

	From Tab.~\ref{table:time} we can see that our pipeline is as efficient as the other methods in OR-Tools. This is because the bulk of the time is invested in the common local search and not in finding the initial solution.

	Lastly, Tab.~\ref{table:tsp} shows our performance on the TSP benchmarks provided by~\cite{vinyals2015pointer}, when we train on \textit{mTSP-train}. Even though our network accomplishes a task which is harder than TSP (mTSP), it also specializes in TSP, and does not neglect it in favor of the more-likely mTSP samples in \textit{mTSP-train}.
	
	\section{Discussion}
	
	The solution we propose to the mTSP is quite general: the input encoding is generic and the output is a straight-forward encoding of the solution. In fact, we directly encode problem specific information only in Alg.~\ref{alg:softassign}. From our experience, the network could train even with one iteration of this algorithm, which is akin to a conventional softmax. However, more iterations are needed, in order to be competitive with all of the sophisticated heuristics of OR-Tools.
	
	Producing an output that is sparse and overparameterized, instead of reduced to the economical sequence representation, is one point in which we differ from the seq2seq approach. It remains to be seen if a sequence generation approach is able to model combinatorial problems with an output space that is more complex than a single permutation.
	
	With regards to other flavors and generalizations of the mTSP, we would like, as future work, to solve the minmax variant by training on such examples and to tackle the VRP, by modifying both the training samples and Alg.~\ref{alg:softassign} to normalize also by the maximal capacity. Solving mTSP with time constraints is a challenge by itself, since these  constraints need to be added to the individual cities' encodings.

	\bibliography{main}
	
\end{document}